\documentclass[english,runningheads,a4paper]{llncs}

\usepackage{xcolor}
\usepackage{tabularx}
\usepackage{amsmath}
\usepackage{amsfonts}
\usepackage{graphicx,newtxtext}
\usepackage[bottom]{footmisc}
\usepackage{comment}
\usepackage{afterpage}

\usepackage{colortbl}

\usepackage{url}
\usepackage{booktabs}
\usepackage{todonotes}
\usepackage{multirow}
\usepackage{subcaption}
 \usepackage{eucal}
 
\spnewtheorem{definition}{Definition}{\bfseries}{\rmfamily}

\begin{document}
%

\title{Semantic-based End-to-End Learning for Typhoon Intensity Prediction}

\author{Hamada M. Zahera\and Mohamed Ahmed Sherif\and
	Axel Ngonga}

\authorrunning{Zahera et al.}

\institute{DICE Group, Department of Computer Science,Paderborn University\\\email{\{hamada.zahera, mohamed.sherif, axel.ngonga\}@uni-paderborn.de}}

\maketitle

\begin{abstract}
Disaster prediction is one of the most critical tasks towards disaster surveillance and preparedness. 
Existing technologies employ different machine learning approaches to predict incoming disasters from historical environmental data. 
However, for short-term disasters (e.g., earthquakes), historical data alone has a limited prediction capability.
Therefore, additional sources of warnings are required for accurate prediction. 
We consider social media as a supplementary source of knowledge in addition to historical environmental data. 
However, social media posts (e.g., tweets) is very informal and contains only limited content. 
To alleviate these limitations, we propose the combination of semantically-enriched word embedding model to represent entities in tweets with their semantic representations computed with the traditional \emph{word2vec}. 
Moreover, we study how the correlation between social media posts and typhoons magnitudes (also called intensities) \textit{-in terms of volume and sentiments of tweets-}.
Based on these insights, we propose an end-to-end based framework that learns from disaster-related tweets and environmental data to improve typhoon intensity prediction. This paper is an extension of our work originally published in K-CAP 2019~\cite{zahera2019jointly}. We extended this paper by building our framework with state-of-the-art deep neural models, updated our dataset with new typhoons and their tweets to-date and benchmark our approach against recent baselines in disaster prediction. 
Our experimental results show that our approach outperforms the accuracy of the state-of-the-art baselines in terms of F1-score with (CNN by $12.1\%$ and BiLSTM by $3.1\%$) improvement compared with last experiments. 
\end{abstract}

\section{Introduction}
Disaster prediction and early warnings are crucial when mitigating the impact of disasters and consequent damage~\cite{glade2014early}. 
Even with the significant improvements in forecasting and warning systems, there are many factors that still limit the accuracy of the current prediction algorithms such as: the lack of complete data on natural hazards, monitoring instruments and the highly dynamic nature of natural hazards~\cite{reese2016we}. 
Interestingly, social media plays an increasingly significant role in disaster management and communication~\cite{reuter2018fifteen}. 
People use social media during disasters to share their feelings, ask for help and provide disaster relief efforts. 

A significant body of research has hence leveraged shared disaster-related information in social media to reduce the impact of disasters and deliver faster responses~\cite{socialmedia2,socialmedia3}. 
For instance, the authors of \cite{sakaki2010earthquake} analyzed user tweets during 25 different earthquakes in Japan, where they demonstrated how social media users can act as a reliable source to provide real-time situational updates during disasters. 
On the other hand, decision makers use social media to engage with the public quickly and widely. 
For example, during typhoon \emph{Pablo} in 2012, local authorities in the Philippines asked people to use the hashtag \texttt{\#pabloph} for getting or sharing on-site updates about the typhoon~\cite{irevolutions_2013}.

Such correlations are valuable for supporting decision makers in emergency response processes.
Previous works used data mining techniques to extract such correlation. 
For instance, \cite{anam2018evaluating} applied a wavelet analysis to track the disaster progression from social media data.
Their results showed that wavelet-based features can preserve text semantics and predict the total duration for localized small-scale disasters.  

In this work, we propose an end-to-end learning model to classify the intensity of typhoons (also called a typhoon's category or class~\cite{chen2012upper}) 
by learning from environmental data and social media (\emph{tweets}). 
We were inspired by previous works (see, e.g.,~\cite{tompson2014joint,qin2016joint}) which suggest that the joint learning of multiple models can significantly outperform standalone models. 
Our proposed approach consists of two jointly-trained models. 
The first model (dubbed \emph{Feature Extractor}) analyzes typhoon-related tweets and computes statistical features (i.e., tweets volume and sentiments variances). 
To capture tweet sentiments, we employ a semantics-enriched word embedding in which \textit{entities} are recognized and represented as semantics vectors.
The second model (dubbed \emph{Typhoon Classifier}) takes an input of the combined features extracted by the first model and the environmental data. 
Both models are trained jointly through a shared loss function and their learning parameters are optimized using the same gradient descent.

We evaluate our joint model in experiments on two real sources of data.  
First, we use environmental data of typhoons tracked by the \emph{Joint Typhoon Warning Center (JTWC)}. 
The dataset contains measurements of climate changes (e.g., wind speed and pressure of sea level) before, during and after typhoon landfall. 
As a second dataset, we rely on typhoon-related tweets collected using keyword-based queries executed during periods of typhoons between 2006-2018. We employ different architectures based on Deep Neural Networks (DNN), Deep Convolutional Network (CNN) and Recurrent Neural Networks (e.g, RNN, LSTM and BiLSTM) as our baseline approaches. 
Our results suggest that our jointly-trained models outperform these baseline solutions in disaster prediction. We summarize the main contributions in this paper as follows:
\begin{enumerate}
	\item We propose a \textit{generic} end-to-end framework that improves the overall system performance via joint learning. 
	\item We conduct several experiments on a real disaster dataset to evaluate the performance of our proposed approaches.
	Our results clearly show that our proposed framework outperforms the state-of-the-art standalone baselines significantly.
	
	\item We studied the impact of incorporating semantics embedding from knowledge graphs to enrich tweets representation. Our experiments show that feeding our model with semantics representation of the entities included in the tweets improved the system overall performance.
	
	\item We provide an updated version our disaster dataset (TED), which includes typhoons environmental data and their tweets up-to 2018 (the last archived date by JTWC).
	
\end{enumerate}
All our implementations are open-source and available from the project website\footnote{\url{https://github.com/dice-group/joint-model-disaster-prediction}}.

\section{Data and Preliminaries}
\label{sec:pre}


In this section, 
we discuss our social media analysis during different typhoons and the dataset collection and preprocessing. 

\subsection{Social Media Content Analysis During Typhoons} 
\label{sec:social_media_analysis}

Typhoon environmental data are tracked periodically (i.e. at regular time intervals) before, during and after they strike.
The goal of our work is to detect the intensity of typhoons not only based on such environmental data but also based on data collected by humans in the form of social media posts during each of the typhoons under study.

\begin{figure}[h!t]
    \centering
    \begin{subfigure}[b]{0.9\linewidth}
        \centering
         \includegraphics[width=\linewidth]{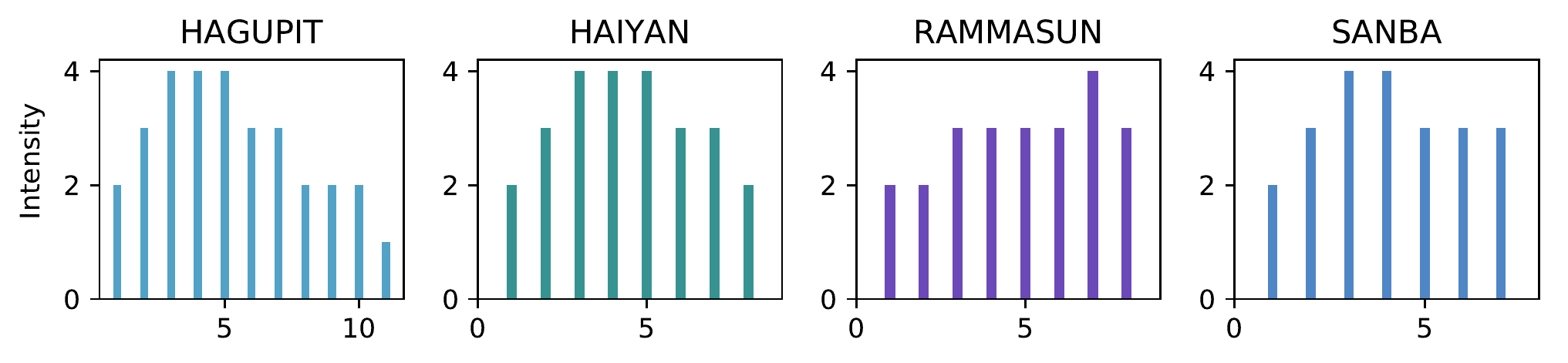}
    \label{fig:typhoon_intensity}
    \end{subfigure}

  \begin{subfigure}[b]{0.9\linewidth}
      \centering
      \includegraphics[width=\linewidth]{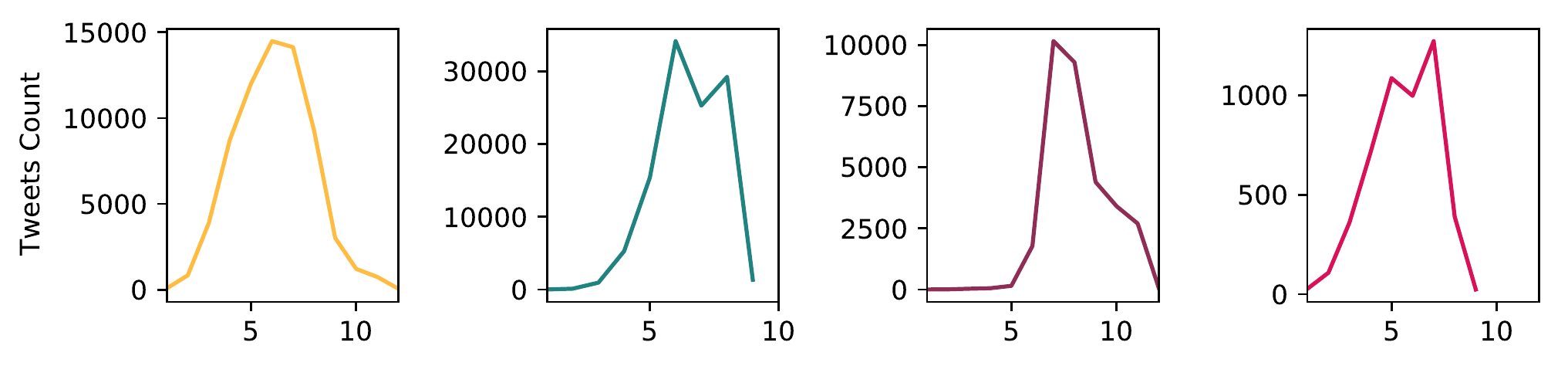}
    \label{fig:tweets volume}
  \end{subfigure}
   
  \begin{subfigure}[b]{0.9\linewidth}
      \centering
      \includegraphics[width=\linewidth]{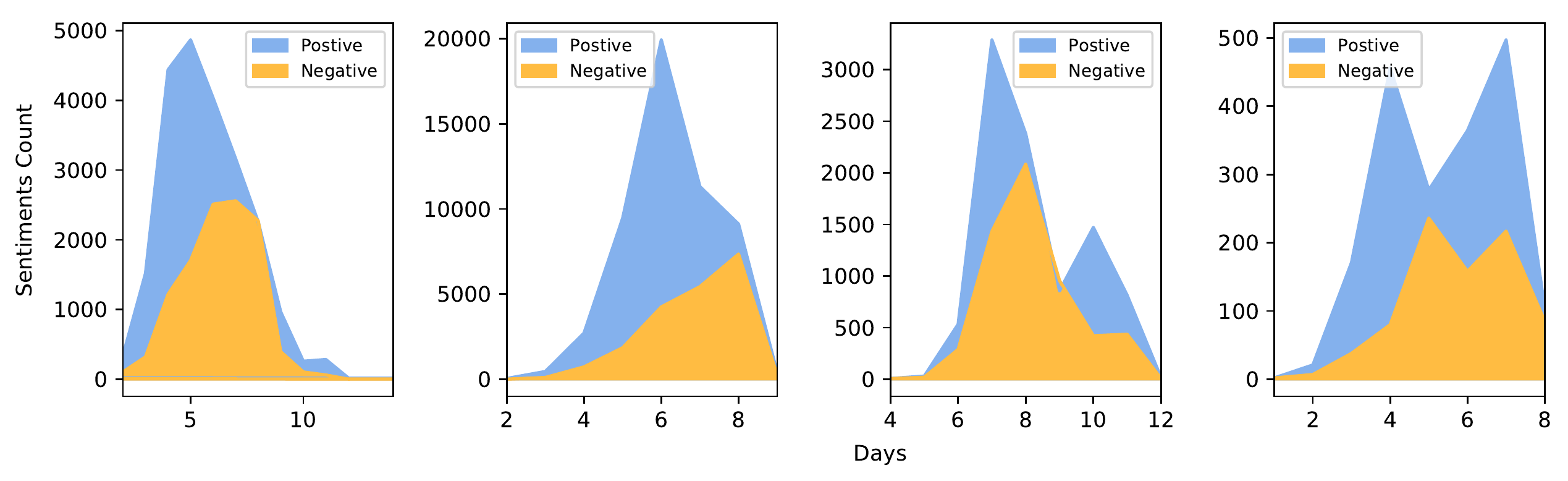}
    \label{fig:tweets sentiment}
  \end{subfigure}

    \caption{Our analysis of social media during the typhoons HAGUIT, HAIYAN, RAMMASUN, and SANBA}
    \label{fig:social_media}
\end{figure}

To pair social media with the environmental data of typhoons, we collect all tweets posted within the time slot of the respective environmental data into one batch. 
Inspired by the work of~\cite{tweetsFeatures1,tweetsFeatures},  
we thus analyzed the volume of tweets as well as sentiments during different time slots of typhoons. 
As shown in Figure~\ref{fig:social_media}, the upper row of plots depicts our content analysis of tweets during four different typhoons,
where we explored how typhoon intensities vary during typhoon days.
In the middle row of plot of the same figure, we present the count of tweets within the same time slots of the provided intensity in the upper row.
Finally, the lower 4 plots present count of tweets with  positive (in blue) and negative (in yellow) sentiment. 
Comparing the respective plots of intensity, tweets count and sentiments for each of the 4 typhoons, we are able to see the correlation between typhoon's intensity and tweets count and sentiment.
To this end, we use the count of tweets and their sentiments as additional indicators for predicting typhoon intensity. 

\subsection{Dataset}~\label{sec:dataset}
We used three datasets to evaluate the performance of our proposed models 
(details of each dataset are presented in Table~\ref{tab:datasets}).

\begin{itemize}
    \item[-] \emph{JTWC Best-tracked ({typhoons environmental data}).} 
    This dataset consists of $3,162$ tracked data points for 70 typhoons between 2006-2016.
    Each data track is labelled into $4$ classes (\texttt{TD}: tropical depression, \texttt{TS}: tropical storm, \texttt{TY}: typhoon and \texttt{ST}: super typhoon).
    The typhoon data also includes time-stamp, location, maximum wind speed (\texttt{VMAX}), wind intensity (\texttt{RAD}) and sea level pressure (\texttt{MSLP}). 
    Note that we removed noisy and corrupted data (e.g., missing values). Then, we handled the problem of imbalance classes using the SMOTE technique~\cite{chawla2002smote} 
    and over-sampled minority classes. 
    \item[-] \emph{Typhoon Tweets.} 
    To search and retrieve typhoon-related tweets, we executed keywords queries based on related typhoon terminology such as \textit{typhoon} and typhoon names (e.g., \textit{Haiyan}). The official Twitter streaming API limits free access\footnote{\url{https://developer.twitter.com/en/docs/tweets/search/overview}} to tweets into only the past 7 days. 
   However, we were able to get all tweets using the open-source library \emph{GetOldTweets-python}\footnote{https://github.com/Jefferson-Henrique/GetOldTweets-python}. 
    \item[-] \emph{Stanford NLP Sentiment140.} We used the Stanford sentiment dataset to enrich our model's performance in sentiment detection. This dataset contains $1.6$ million labeled tweets with binary sentiments (positive/negative).
    
\end{itemize}
\begin{table}[h!tb]
\caption{Details of datasets.\label{tab:datasets}}
	\centering
		\begin{tabularx}{0.75\linewidth}{@{}p{.25\textwidth}p{.15\textwidth}p{.15\textwidth}p{.15\textwidth}}
			\toprule 
			\textbf{Dataset} & \textbf{Training} & \textbf{Testing} & \textbf{Classes} 
			\\ 
			\midrule
			JWTC Best-Track & 2,529 & 633 & 4 
			\\ 
			Typhoon Tweets & 1,052,599 & 270,364 & unlabeled 
			\\ 
			Sentiment140 & 1,280,000 & 320,000 & 2 
			\\ 
			\bottomrule
		\end{tabularx}

\end{table}
\paragraph{Data Preprocessing} 
\label{sec:tweets_rep}
Tweets are commonly informal and often contain noisy and incomplete text. Hence, the preprocessing of tweets often involves varied techniques to achieve high-quality analysis in data mining applications. 
In particular, we carry out the following preprocessing steps: 
\begin{itemize}
    \item[-] \emph{Cleaning up.} We remove URLs, non-ASCII characters, usernames and hashtags from all tweets. It is a common process to remove stop words in standard text preprocessing. However, in our work, we keep stop words to preserve the context of words and obtain an accurate sentiment analysis~\cite{saif2012semantic}.
    \item[-] \emph{Entity recognition.} There are several tools for entity extractions and semantics reasoning.~We use the Spacy API\footnote{\url{https://spacy.io/}} to annotate entities from tweets. Spacy is an open-source Natural Language Processing (NLP) library that is widely used due to its availability and preeminent accuracy in different linguistics tasks~\cite{al2017choosing}.
    \item[-] \emph{Tokenization.} Our final tweet preprocessing step is to tokenize tweets into words, then convert to lower-case letters. 
\end{itemize}

 \begin{example}
  The preprocessing of the tweet \texttt{"My heart goes out to all those affected by Typhoon Haiyan. You can help by donating to the Phi\-l\-ip\-pi\-ne RED CROSS here (link: \url{http://www.redcross.org}) \url{redcross.org}"} will remove the URL at the end of the tweet and generate the following tokenized word vector:
 [\texttt{my, heart, goes, out, to, all, those, affected, by, {typhoon}, {haiyan}, you, can, help, by, donating, to, the, {philippine}, {red}, {cross}, here}].
 \end{example}

\section{The Approach} 
\label{sec:approach}
\begin{figure*}[h!tb]
	\centering
	\captionsetup{justification=centering}
	\includegraphics[width=0.70\linewidth]{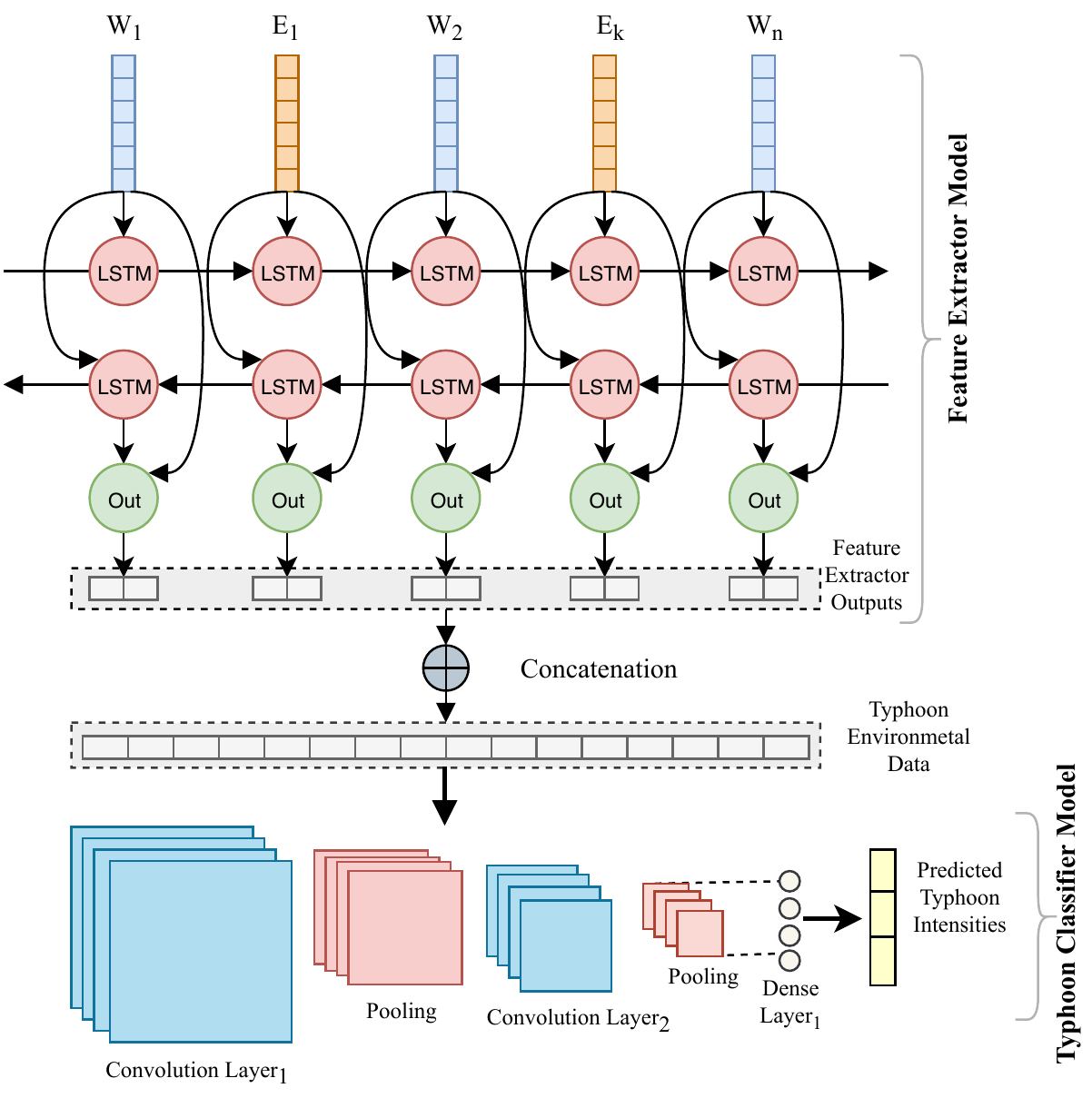}
	\caption{Our Joint Model (BiLSTM+CNN) for typhoon intensity classification. 
	The entities ($e_k$) vectors (in orange) are extracted from knowledge graph (e.g., ConceptNet). 
	The words ($w_n$) vectors (in blue) are obtained from our word embedding. 
}
	\label{fig:jointtraining}
\end{figure*}

In this section, first we discuss the problem formulation of jointly training from social media and environmental data to improve typhoon intensities. 
Our approach takes two inputs: 
typhoon environmental data and tweet batches. 
Figure~\ref{fig:jointtraining} shows the architectures (BiLSTM+CNN) for our jointly models (Feature Extractor and Typhoon Classifier). 
In the rest of this section, we discuss our semantics-enriched word embedding to represent tweets and joint models architectures.
\subsection{Problem Formulation}
\label{sec:joint_training}
Let $D=\{\langle x_1,y_1\rangle \dots \langle x_n,y_n\rangle \}$ be a typhoon environmental data, where $X = \{ x_1 \dots x_n\}$ is a set of $n$ typhoon observations and $Y = \{ y_1 \dots y_j\}$ is a set of typhoon categories (i.e., labels, classes). 
Each $x_i \in X$ represents an instance of a typhoon data with $m$ features (e.g., time-tamp, wind-speed, sea level pressure and gust) and each $y_i \in Y$ represents the respective typhoon's category (e.g., either a tropical-depression, tropical-storm, typhoon or super-typhoon). 

For each typhoon with environmental data $x_{i}$, we collect all related tweets posted within the time-stamp of $x_{i}$, we dub such tweets as $T$. 
Further, we analyse all tweets to extract statistical features (i.e., tweets-based features) such as \textit{tweets volume} and \textit{variances of tweets'  sentiments}. 
Finally, we combine these features with typhoon's environmental data in one input vector.
\begin{definition} {\textbf{(Task Description.)}}
Our goal is to design a classification model able to learn features from $D$ and $T$ in order to predict the typhoon category ($Y$). 
We build our classification model as a joint model of two cascaded models $F_1$ and $F_2$ dubbed \textit{feature extractor} and \textit{typhoon classifier} models respectively (i.e., $F_1 (T)+F_2 (D)\longrightarrow Y$). 
To ensure the joint training of $F_1$ and $F_2$, we combine the loss functions of both models ($L_{F_1}$,$L_{F_2}$) in one shared loss function as follows:
\begin{equation}\label{eq:joint}
L_{joint}=\lambda_{F_1} \cdot L_{F_1}+ \lambda_{F_2} \cdot L_{F_2}
\end{equation}

Where $\lambda$ parameter is used to balance the individual loss functions ($L_{F_1}$, $L_{F_2}$). In this paper, we set all $\lambda$ parameters as 1. To compute the training losses, we used the cross-entropy function as a loss function as follows: 
\begin{equation}\label{equ:cross-entropy}
L=-\frac{1}{n}\sum_{i=1}^{n}\left [y_i\log(H(x_{i})))+(1-y_i)\log(1-H(x_{i}))\right]
\end{equation}
where $y_i$ and $H(x_{i})$ donate target and predicted typhoon categories respectively for typhoon instance $x_i$. 
\end{definition}

\begin{definition}{\textbf{(Joint-Learning).}} 
Let $\phi_1$ and $\phi_2$ be the learning parameters of \textit{feature extractor} and \textit{typhoon classifier} models. Both $\phi_1$ , $\phi_2$ are optimized concurrently as follows:
Assume two consecutive batches of training data $B_{t}$ and $B_{t+1}$, the learning parameters 
in $B_{t}$ are optimized using the same gradient descent (e.g., ADAM optimizer) 
by backpropagating the gradients to both models.  
In the following batch ($B_{t+1}$), the computation of statistical features by \textit{feature extractor} model is hence further \textit{adapted} not only from the losses in its outputs, but also from the losses in the final output by \textit{typhoon classifier} model.
Therefore, the \textit{feature extractor} model feeds adaptive features from social media to the \textit{typhoon classifier} model.
\end{definition}

\subsection{Semantics-enriched Word Embedding}
\label{sec:semantic_embedding}
Our analysis of tweets during different typhoons (see Section~\ref{sec:social_media_analysis}) suggests that we can use the tweets volume and sentiments as additional features in our model.
To perform sentiment analysis on crisis tweets, we used the \emph{continuous skip-gram} approach~\cite{mikolov2013distributed}
to train our word embedding model on typhoon tweets and the \emph{sentiment140} dataset. 
While generic word embeddings trained on generic large-scale datasets (e.g., Wikipedia and Google News) could have been used here, they often do not capture domain-specific knowledge and semantic nuances.
In contrast, domain-adapted word embeddings are effective in the field of the context in which they are trained as they capture domain-specific knowledge~\cite{sarma2018domain}.


Now, given a preprocessed list of words of a tweet $t=(w_{1},w_{2},e_{1},\dots,e_{j},w_{s})$, we map each word $w_{i}$ to its embedding vector in $\mathbb{R}^{1 \times d}$ with $d$ dimension. 
Unlike classical word embeddings, we represent entities $(e_{1},\dots,e_{j})$  with their corresponding vectors from the semantic knowledge base of \emph{ConceptNet}\footnote{\url{http://conceptnet.io/}}, where entities and relationships are projected into the same embedding space. 
The motivation behind combining semantics embedding with the classical word embedding is the superior performance of semantics embedding in modern data mining applications~\cite{paulheim2018make}. 

 \begin{example}
  The semantics-enriched word embedding vector of the tweet from Example~1 would be 
 [\texttt{my, heart, goes, out, to, all, those, affected, by, \textbf{typhoon}, \textbf{haiyan}, you, can, help, by, donating, to, the, \textbf{philippine},} \texttt{\textbf{red}}, \texttt{\textbf{cross}}, \texttt{here}]. Note that, the bold words represent semantic entities.
 \end{example}

For each input tweet, we build an embedding matrix $M \in \mathbb{R}^{s\times|d|}$, where $s$ is the number of words per tweet. Each row $i$ of $M$ represents the \textit{word2vec} embedding of the $w_i$ at the corresponding position $i$ in a tweet.
Our \textit{word2vec} model has a dimension $d$ of $200$ and vocabulary size of $47,137$ words and $1,152$ recognized entities.
Due to variable lengths of tweets, we fixed $s$ to the average number of words per tweet to maintain a regular embedding matrix. 
For this reason, we truncated longer tweets and padded shorter tweets with zeros.

\subsection{Feature Extractor}
\emph{Feature extractor} is the first model in our approach that aims to extract statistical features from tweets. To model the words sequences in tweets, we employ a bidirectional-LSTM (BiLSTM) model. 
First, we use an embedding look-up layer to map words to their corresponding vectors from semantics-enriched word embedding model. 
Then, we employ on BiLSTM layer with $64$ units and dropout rate $0.25$, and one dense layer with softmax output. 

Given a sequence of input words $(w_{1},w_{2}, \dots,w_{s})$, BiLSTM see the context of word $w_s$ in both directions (left-to-right $\overrightarrow{h}$ and right-to-left $\overleftarrow{h}$) and summarizes information into a concatenated output vector $[\overrightarrow{h} \boldsymbol{\cdot} \overleftarrow{h}]$. 
BiLSTM layer associates each time-stamp with an input $i_{t}$, memory cell $m_{t}$, forget gate $f_{t}$ and an output gate $o_{t}$. The output vector $h_{t}$ is then computed by iterating the following equations:
\begin{equation}
\begin{split}
    \boldsymbol{i_{(t)}}&=\sigma(\theta_{xi}^T.x_{(t)}+\theta_{hi}^T.h_{(t-1)}+b_{i})
    \\
    \boldsymbol{f_{(t)}}&=\sigma(\theta_{xf}^T.x_{(t)}+\theta_{hf}^T.h_{(t-1)}+b_{f}) 
    \\
    \boldsymbol{o_{(t)}}&=\sigma(\theta_{xo}^T.x_{(t)}+\theta_{ho}^T.h_{t-1}+b_{o})
    \\
    \boldsymbol{g_{(t)}}&=\tanh(\theta_{xg}^T.x_{(t)}+\theta_{hg}^T.h_{(t-1)}+b_{g})
    \\
    \boldsymbol{m_{(t)}}&=f_{(t)}\otimes m_{(t-1)}+i_{(t)}\otimes g_{(t)}
    \\
    \boldsymbol{h_{(t)}}&=o_{(t)}\otimes \tanh(m_{(t)})
\end{split}
\end{equation}

Where $\theta_{xi},\theta_{xf},\theta_{xo},\theta_{xg}$ are the weights vectors of the \textit{input, forget, memory and output} gates 
concerned with the input vector $x_{(t)}$. 
Respectively, $\theta_{hi},\theta_{hf},\theta_{ho},\theta_{hg}$ are the weights vectors concerned with the previous hidden vector $h_{(t-1)}$.
$b_{i},b_{f},b_{o},b_{g}$ are the bias terms for the four gates. $\sigma$ and $\otimes$ donates \emph{sigmoid} function and \emph{product-wise} multiplication respectively.
 
The outputs of the BiLSTM are probabilities of positive and negative sentiments computed by softmax function in Equation~\ref{eq:softmax}. 
Subsequently, we extract and combine statistical features from BiLSTM outputs with typhoon data as $D \in~ \mathbb{R}^{n\times [m+c,v_{-},v_{+}]}$ with $n$ typhoon instances and $m$ features. $c,v_{-},v_{+}$ represents the statistical features: \textit{tweets count, variance of negative sentiments and variance of positive sentiments}.
The variances of sentiments are computed as follows:
\begin{equation}\label{eq:variance}
    v = \frac{1} {c} \displaystyle\sum_{i=1}^{c}(\mathcal{S}_i - \mu)^2
\end{equation}
Where $\mathcal{S}_i$ donates the predicted sentiment of tweet $i$ , $\mu$ is the average of sentiments and $c$ is the tweets count.

\begin{equation}\label{eq:softmax}
P(y=j|z)=\frac{e^{z_{j}}}{\sum_{i=1}^{k} e^{z_{k}}}
\end{equation}
Given input vector $z=\theta^Tx$ and $k$ is the number of typhoon categories, $P(y=j|z)$ donates the probability of typhoon category $j$.

\subsection{Typhoon Classifier}~\label{sec:Typhoonclassifier}
The \emph{Typhoon classifier} takes input features from the \textit{Feature Extractor} (see previous Section) to predict the typhoon intensity as a final output. 
We explored different deep architectures models (i.e., the DNN, CNN and RNN models) as baselines to benchmark the performance of our proposed joint-training model for typhoon predictions. In the following, we discuss the architecture of \emph{Typhoon Classifier} presented in our joint approach in Figure~\ref{fig:jointtraining}.

We employed two convolutional layers (with ReLU activation function). The first CNN layer defines a filter (or also called feature detector) of $32$ output dimension and kernel size $3$. Only defining one filter would allow the neural network to learn one single feature in the first layer. 
The result from the first CNN layer is fed into the second CNN layer, where another filter is defined with $16$ output dimension and kernal size $3$. 
After each convolutional operation, we subsample the output by Max-pooling layer (also called pooling operation). 
We use a max-pooling layer to eliminate non-maximal values and reduces computation in later layers. 
We also employed dropout rates in convolutional layers ($0.30$ after first layer and $0.2$ after second layer), 
to avoid overfitting and increase the model robustness. 
Finally, we used a fully connected layer with softmax activation function to compute the output probabilities for all typhoon categories as in Equation~\ref{eq:softmax}. 
At the end, the class with the highest probability is returned as the model output.



\section{Experiments} 
\label{results}
We conducted several sets of experiments to benchmark the performance of the baseline models as well as our proposed approach to predict the intensity of typhoons.  
The aim of our evaluation is to answer the following two research questions: 

\begin{itemize}
    \item[$Q_1.$] To which extent, can social media improve the performance of the state-of-the-art disaster prediction approaches?
    \item[$Q_2.$] What is the impact of semantic embedding of tweets representation on the performance of our proposed approach? 
\end{itemize}

In the rest of this section, we begin by describing baseline approaches and evaluation metrics. 
Thereafter, we analyze our results and answer each our research questions in details.

\subsection{Baselines}
We benchmarked our approach against different baselines including traditional ML and deep neural models.
We selected the SVM classifier as our traditional baseline classifier as it outperform the other traditional classifiers~\cite{burel2017semantics}.
For our experiment, we implemented an RBF-based kernel SVM classifier trained with typhoon environmental data.
We also used four benchmarks from the deep neural based models (i.e., DNN and RNN, CNN and BiLSTM),
where both achieved good performance in disaster-related research~\cite{chen2019detecting,wang2017using}. 
In particular, CNN-based model with semantics has been employed to classify disaster-related social media data. 
The authors of~\cite{burel2017semantics} suggested to enrich data representation by recognizing entities (0 or 1) in tweets (also called bag of concepts) and add as additional features. However, their does not capture the context of an entity where it exist. 

In our approach, we recognized entities and extract their semantic representation from an external knowledge graph embedding. The semantics embedding helps to consider not only the existence of an entity, but also represent the context of an entity where it exist.
We specify our proposed approaches used in the experiments as follows:
\begin{itemize}
    \item \textbf{LSTM+DNN (word embedding)}: Our first approach of two deep neural models (LSTM and DNN) are jointly trained with combined features from environmental data and word embedding.
    \item \textbf{LSTM+DNN (semantic embedding)}: This model is the same as LSTM+DNN (word embedding), but we consider semantic-enriched word embedding for data representation. See Section~\ref{sec:semantic_embedding} for more details.
    \item \textbf{LSTM+RNN (word embedding)}: Our second approach of two deep neural models (LSTM and RNN) trained with combined features from environmental data and word embedding.
    \item  \textbf{LSTM+RNN (semantic embedding)}: This model is the same as LSTM+RNN (word embedding), with considering semantic-enriched word embedding.
    \item \textbf{BiLSTM+CNN (word embedding)}: Our third approach of two deep neural models (BiLSTM and CNN) are jointly trained with combined features from environmental data and word embeddings.
    
     \item \textbf{BiLSTM+CNN (semantic embedding)}: This model is the same as BiLSTM+CNN (word embedding) with considering semantic-enriched word embedding. 
    
\end{itemize}
To ensure a fair performance evaluation, we evaluated our proposed  models with the same architectures and hyper-parameters settings used to configure the baselines. 
Moreover, we evaluated the impact of incorporating semantics embedding from external knowledge graphs in comparison with the traditional word embedding of our input tweet dataset.

\subsection{Evaluation Metrics} 
We considered standard evaluation metrics to assess the models performance in the task of typhoons prediction. 
We divided the dataset (formally described in Section~\ref{sec:dataset}) into train-test splits of 80\%-20\% respectively.
We train each model for $100$ epochs, then we evaluated the overall performance metrics (accuracy, precision, recall and F1-score) on the test dataset as depicted in Table~\ref{tab:performance}.

\begin{table}[htb]
\centering
\caption{Performance evaluation on test dataset using \emph{Accuracy} (A), \emph{Precision} (P), \emph{Recall} (R) and \emph{F1-Score} (F). Our state-of-the-art models and their baselines are marked in gray.\label{tab:performance}}
\begin{tabularx}{0.88\linewidth}{@{}p{.2\columnwidth}p{.22\columnwidth}p{.1\columnwidth}p{.1\columnwidth}p{.1\columnwidth}rp{.1\columnwidth}}
\toprule
\textbf{Model} & \textbf{Description} & \textbf{A} & \textbf{P} & \textbf{R} & \textbf{F}  \\  \hline \midrule
  SVM & Baseline & 0.579 & 0.347 & 0.579 & 0.430\\
  DNN & Baseline &   0.756  &   0.809    &      0.756   &  0.781  
  \\
  RNN & Baseline &  0.802   &  0.827   &   0.802       &    0.814  \\
\rowcolor{gray!20}
 CNN & Baseline &  0.702 & 0.918 & 0.702 &  0.796 \\
\rowcolor{gray!20}
 BiLSTM & Baseline & 0.840 & 0.880 & 0.840 & 0.859 \\
  \midrule
  LSTM+DNN & Word emb. &   0.873 & 0.892  & 0.873  & 0.882    \\
  LSTM+DNN & Semantic emb. & 0.917& 0.922 & 0.925 & 0.917 \\
  LSTM+RNN & Word emb. & 0.860  & 0.875   & 0.860 &  0.855    \\
  LSTM+RNN & Semantic emb. &  0.891 &  0.904 & 0.891  &  0.891  \\
  \rowcolor{gray!20}
 BiLSTM+CNN & Word emb. & 0.847 &  0.938 & 0.847 & 0.890 \\
  \rowcolor{gray!20}
 BiLSTM+CNN & Semantic emb. & 0.902 & 0.933 & 0.902 & 0.917 \\
  \bottomrule
\end{tabularx}
\end{table}
\subsection{Discussion and Result Analysis}

\paragraph{\textbf{To answer $Q_1$,}} we evaluated the baseline models with features extracted from the environmental data (see Section~\ref{sec:dataset}). 
Further, we used the same features, in addition to the features extracted from relevant tweets (i.e., tweets-based features) to train our proposed models.
In particular, we computed the additional features of tweets count $c$, variance of positive $v_{+}$ and negative $v_{-}$ sentiment of tweets.


Table~\ref{tab:performance} shows how challenging is the task of typhoons intensity prediction, where the best baseline model (i.e., the BiLSTM model) produces an accuracy of $0.84$. 
All the base line models were trained on environmental data, which captured from sensor devices and usually include noisy and incomplete data~\cite{morton2011challenges}. 
In contrast, our proposed models clearly demonstrate a significant improved performance where tweets-based features were incorporated with environmental data. 
In particular, the best accuracy is achieved by our proposed model BiLSTM+CNN, where it outperforms the respective baselines (CNN by $12.1\%$ and BiLSTM by $3.1\%$) on micro average F1-score. The other proposed models (LSTM+DNN and LSTM+RNN) also outperform their respective baselines (by $11\%$ in DNN and $5.8\%$ in RNN) on average. 

\begin{figure}[t!]
	\centering
	\includegraphics[width=0.7\linewidth]{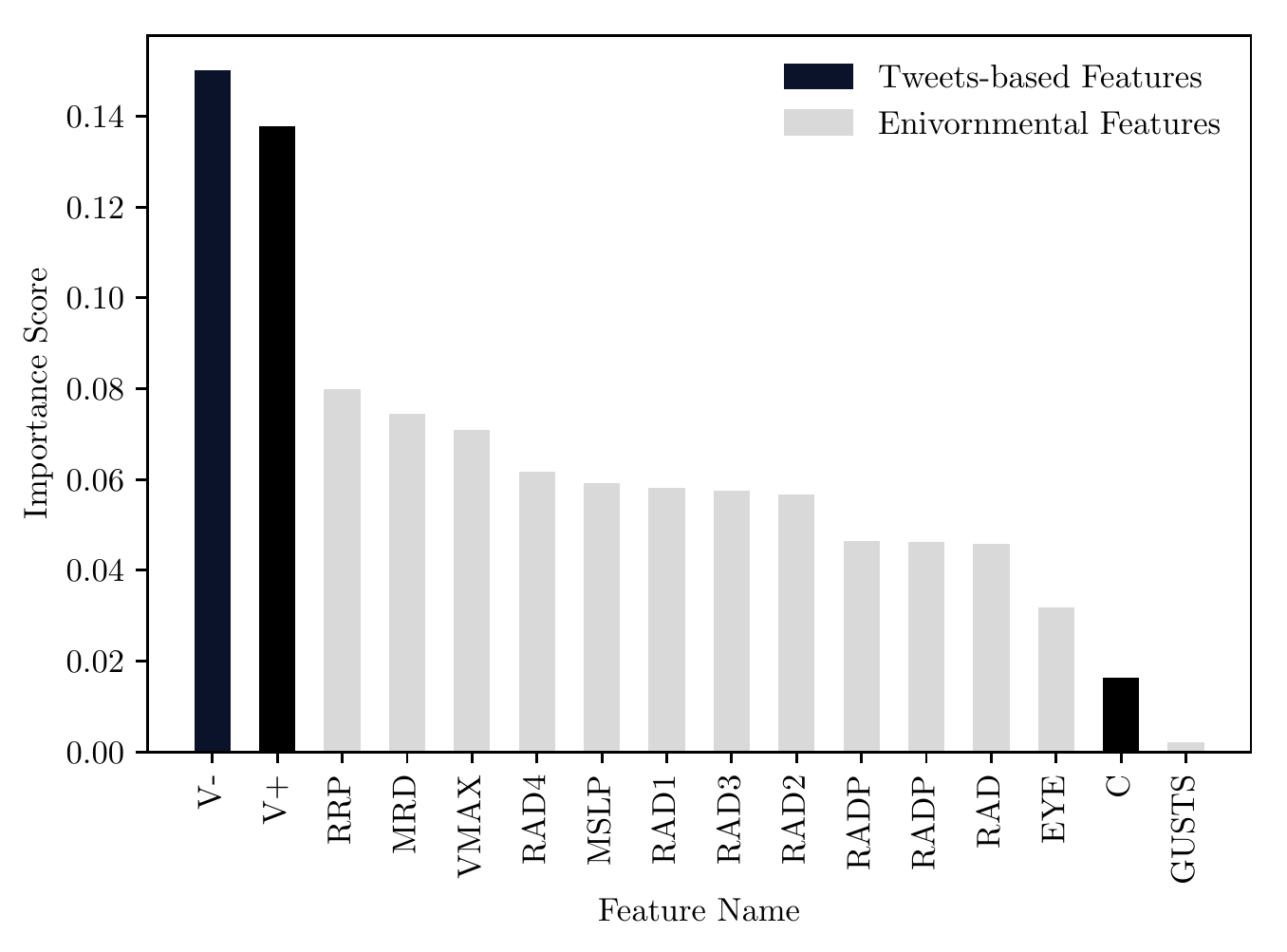}
	\caption{Importance of environmental and tweets-based features.}
	\label{fig:feature-importance}
\end{figure}

To summarize our answer, training our proposed model with relevant features from social media in addition to environmental data outperforms both the accuracy and the F1-measure of the baseline approaches. 
To understand the superior performance of our models, we evaluated the importance of training features using the \emph{Random Forests} algorithm\footnote{We used the implementation from scikit-learn \url{http://scikit-learn.org/stable/modules/generated/sklearn.ensemble.RandomForestClassifier.html}}.
As shown in Figure~\ref{fig:feature-importance}, tweet-based features were found to be more important than environmental features. As discussed in Section~\ref{sec:joint_training}, tweet-based features are \textit{adaptive} to prediction losses in our joint model which helps to fit the features by joint training and improve the accuracy of final prediction.

\paragraph{\textbf{To answer $Q_2$,}} we investigated the impact of incorporating semantics embedding on the performance of our proposed system. 
Our experiments showed improved performance in terms of accuracy, precision, recall and F1-measures with the semantics embedding based models over those based only on word embedding. 
In particular, the accuracy of our proposed joint-model architectures are improved by up to $3\%$ in both LSTM+DNN and LSTM+RNN models compared to word embedding based models. 
As discussed earlier in Section~\ref{sec:semantic_embedding}, the semantics embedding from knowledge enrich the representation of entities and their relationship into semantic vectors.

\begin{figure}[h!]
    \centering
    \includegraphics[width=0.80\linewidth]{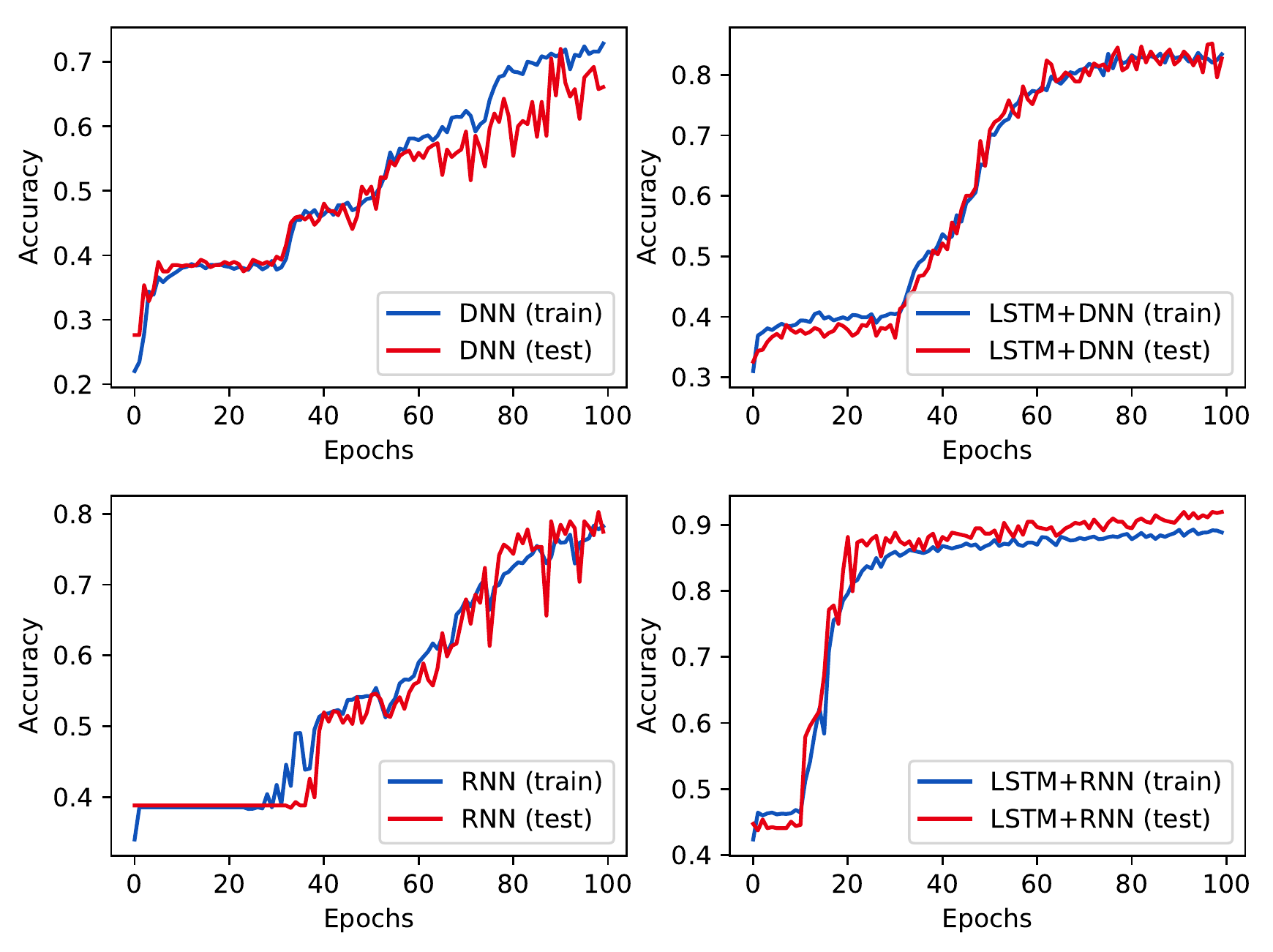}
    \caption{Over-fitting evaluation.}
    \label{fig:overfitting}
\end{figure}

\paragraph{\textbf{Training Robustness.}}
We validated the training robustness in each model and checked the training over-fitting. 
As shown in Figure~\ref{fig:overfitting}, we trained all models to achieve robust performances in testing phase as well as in training and alleviate the over-fitting by tuning the hyper-parameters properly.

\section{Related Works}\label{sec:relatedwork}
Our work is related to social media analysis for disaster management, jointly-trained models and semantics-enrichment data mining. 
In the following, we briefly present the related state of the art in each of these areas in turn.

\subsection{Social media analysis for disaster management}

Several studies leveraged the \emph{role of social media in disasters management}\cite{acar2011twitter,sakaki2010earthquake,varga2013aid,qu2011microblogging}. %
For example, \textit{Yury~et~al.}~\cite{tweetsFeatures1} proposed a social media-based framework to estimate damages initially from social media (i.e., tweets and images). 
Their system  analyzed users activities on Twitter before, during and after the hurricane \emph{Sandy}.
Their results showed a strong correlation between activities on social media and hurricane path. 
Similarly, in this paper, we studied social media behaviors during different typhoons, our analysis showed a an implicit correlation based on tweets volume and sentiments.

Social media has also been shown to be a rapid event detector or so-called~\emph{social-sensing} from the crowd. 
For example, \textit{Takeshi~et~al.}~\cite{sakaki2010earthquake} proposed a probabilistic \emph{spatio-temporal} approach to predict the center and path of natural hazards based on geo-based tweets.
The authors analyzed the behaviors of social media users during different cases of earthquakes and typhoons.
Their experimental results demonstrated a strong correlation between user's behavior on Twitter and natural hazards.
The authors used such correlation as an event detector for earthquakes. 

Although previous works have explored the role of social media (e.g., tweets) in crisis events detection and enhancing situational awareness, few research studies showed how tweets sentiments have been used as discriminative features to improve events prediction. 
For example, \textit{Jingrui~et~al.}~\cite{tweetsFeatures} proposed an optimization framework to improve traffic-jam situations. 
Their system extracted traffic indicators from tweets semantics to improve traffic-jam production.

\subsection{Semantics-enriched data mining}
Researchers have leveraged semantics nuances to boost the performance in machine learning (ML) and data mining tasks. 
The authors of~\cite{burel2017semantics,wang2014concept} extended the traditional bag-of-word models with a bag-of-concepts model extracted from semantics knowledge graphs (e.g. \emph{WordNet}, \emph{DBpedia}). However, the authors represented the presence of concepts as vector of indices within a concept space. 
On the other hand, \textit{Jin Wang~et~al.}~\cite{wang2017combining} leverage information from knowledge graph for short text classification. Their approach associated each short text with relevant concepts. Then, words and concepts were combined to generate its embedding from a pre-traind word embedding model. 

In contrast, in our approach, we leverage semantics embedding from knowledge graph which projects concepts (i.e., entities) and their relationship for conceptualized data representation. 

\subsection{Joint-Learning models}
Recently, joint learning models have achieved superior performances in complex tasks. 
For instance, \textit{Jishnu~et~al}.~\cite{ray2019keyphrase} proposed a jointly-training RNN-based models to extract keyphrases from disaster-related tweets.
Through intersection of two RNN models via joint learning, the experimental results clearly demonstrated a significant performances in comparison with existing baseline approaches. 
In a related task of image classification, \textit{Wanli~et~al.}~\cite{ouyang2013joint} proposed a unified deep model that jointly learn from different components in image classification task. 
Similarly, \textit{Zheng~et~al}.~\cite{zheng2017joint} demonstrated that the joint learning of two deep models could not only separately learn user and item latent factors from review text, but also cooperate with each other to boost the performance of rating prediction. 

Inspired by these works, we propose our joint training model to learn features from social media and environmental data in comparison to individual or ensemble models employed in event detection (i.e. prediction)~\cite{burel2017semantics,pouyanfar2016semantic}
\section{Conclusion}
\label{conclusion}
In this paper, we propose an end-to-end training framework that learns from social media and environmental data to improve disasters prediction. 
In particular, we analyzed typhoon-related tweets to capture additional indicators of typhoon events.
Unlike previous works, we extract \emph{adaptive} features based on joint training models (e.g. BiLSTM+CNN).
The first model (BiLSTM) acts as a \emph{feature extractor} from social media and feeds the second model (CNN) with combined features from tweets and environmental data. 
Furthermore, we study the impact of applying semantically-enriched data representation on the performance of our system.
We employed semantics embedding from the external knowledge graph of \emph{ConceptNet}. 
We conducted several experiments to benchmark the prediction performance of our approach and several baselines. 
Our evaluation showed significantly improved accuracies in our approaches (LSTM+DNN: 87.3\%, LSTM+RNN: 86.0\% and BiLSTM+CNN:0.90\%) when compared to state-of-the-art baselines (DNN: 75.6\%, RNN: 80.2\%, CNN: 0.70\% and BiLSTM: 0.84\%). 
Moreover, feeding our proposed joint models with the semantics representation of entities improved F1-score even further (up to 3\% in the case of LSTM+DNN, up to 4\% in the case of LSTM+RNN and up to 2.7\% in the case of BiLSTM+CNN).

In future work, we aim to construct a domain specific knowledge graph from disaster-related tweets for disaster relief tasks (e.g  event summarization, identifying actionable information). 
In addition, we will carry out a comprehensive study of the effect of generic word embedding models (e.g., Word2vec, Glove) in comparison with embedding from domain-specific corpus, knowledge graphs and conceptualized embedding (e.g., BERT). 
\bibliographystyle{abbrv}
\bibliography{references}

\begin{thebibliography}{10}

\bibitem{irevolutions_2013}
To tweet or not to tweet during a disaster?, Jun 2013.

\bibitem{acar2011twitter}
A.~Acar and Y.~Muraki.
\newblock Twitter for crisis communication: lessons learned from japan's
  tsunami disaster.
\newblock {\em International Journal of Web Based Communities}, 7(3):392--402,
  2011.

\bibitem{al2017choosing}
F.~N.~A. Al~Omran and C.~Treude.
\newblock Choosing an {NLP} library for analyzing software documentation: a
  systematic literature review and a series of experiments.
\newblock In {\em Proceedings of the 14th International Conference on Mining
  Software Repositories}, 2017.

\bibitem{anam2018evaluating}
A.~Anam, A.~Gangopadhyay, and N.~Roy.
\newblock Evaluating disaster time-line from social media with wavelet
  analysis.
\newblock In {\em 2018 IEEE International Conference on Smart Computing
  (SMARTCOMP)}, pages 41--48. IEEE, 2018.

\bibitem{burel2017semantics}
G.~Burel, H.~Saif, M.~Fernandez, and H.~Alani.
\newblock On semantics and deep learning for event detection in crisis
  situations.
\newblock In {\em Proceedings of the International Workshop on Semantic Deep
  Learning (SemDeep)}. ESWC 2017, 2017.

\bibitem{chawla2002smote}
N.~V. Chawla, K.~W. Bowyer, L.~O. Hall, and W.~P. Kegelmeyer.
\newblock Smote: synthetic minority over-sampling technique.
\newblock {\em Journal of artificial intelligence research}, 16:321--357, 2002.

\bibitem{chen2012upper}
X.~Chen, D.~Pan, X.~He, Y.~Bai, and D.~Wang.
\newblock Upper ocean responses to category 5 typhoon megi in the western north
  pacific.
\newblock {\em Acta Oceanologica Sinica}, 31(1):51--58, 2012.

\bibitem{chen2019detecting}
X.~Chen, L.~Zou, and B.~Zhao.
\newblock Detecting climate change deniers on twitter using a deep neural
  network.
\newblock In {\em Proceedings of the 2019 11th International Conference on
  Machine Learning and Computing}, pages 204--210. ACM, 2019.

\bibitem{glade2014early}
T.~Glade and F.~Nadim.
\newblock Early warning systems for natural hazards and risks, 2014.

\bibitem{tweetsFeatures}
J.~He, W.~Shen, P.~Divakaruni, L.~Wynter, and R.~Lawrence.
\newblock Improving traffic prediction with tweet semantics.
\newblock In {\em Proceedings of the Twenty-Third International Joint
  Conference on Artificial Intelligence}, IJCAI '13, pages 1387--1393. AAAI
  Press, 2013.

\bibitem{socialmedia3}
J.~B. Houston, J.~Hawthorne, M.~F. Perreault, E.~H. Park, M.~Goldstein~Hode,
  M.~R. Halliwell, S.~E. Turner~McGowen, R.~Davis, S.~Vaid, J.~A. McElderry,
  et~al.
\newblock Social media and disasters: a functional framework for social media
  use in disaster planning, response, and research.
\newblock {\em Disasters}, 39(1):1--22, 2015.

\bibitem{tweetsFeatures1}
Y.~Kryvasheyeu, H.~Chen, N.~Obradovich, E.~Moro, P.~Van~Hentenryck, J.~Fowler,
  and M.~Cebrian.
\newblock Rapid assessment of disaster damage using social media activity.
\newblock {\em Science advances}, 2(3):e1500779, 2016.

\bibitem{socialmedia2}
P.~M. Landwehr and K.~M. Carley.
\newblock Social media in disaster relief.
\newblock In {\em Data mining and knowledge discovery for big data}, pages
  225--257. Springer, 2014.

\bibitem{mikolov2013distributed}
T.~Mikolov, I.~Sutskever, K.~Chen, G.~S. Corrado, and J.~Dean.
\newblock Distributed representations of words and phrases and their
  compositionality.
\newblock In {\em Advances in neural information processing systems}, pages
  3111--3119, 2013.

\bibitem{morton2011challenges}
M.~Morton and J.~L. Levy.
\newblock Challenges in disaster data collection during recent disasters.
\newblock {\em Prehospital and disaster medicine}, 26(3):196--201, 2011.

\bibitem{ouyang2013joint}
W.~Ouyang and X.~Wang.
\newblock Joint deep learning for pedestrian detection.
\newblock In {\em Proceedings of the IEEE International Conference on Computer
  Vision}, 2013.

\bibitem{paulheim2018make}
H.~Paulheim.
\newblock Make embeddings semantic again!
\newblock In {\em International Semantic Web Conference
  (P\&D/Industry/BlueSky)}, 2018.

\bibitem{pouyanfar2016semantic}
S.~Pouyanfar and S.-C. Chen.
\newblock Semantic event detection using ensemble deep learning.
\newblock In {\em Multimedia (ISM), 2016 IEEE International Symposium on},
  pages 203--208. IEEE, 2016.

\bibitem{qin2016joint}
H.~Qin, J.~Yan, X.~Li, and X.~Hu.
\newblock Joint training of cascaded cnn for face detection.
\newblock In {\em Proceedings of the IEEE Conference on Computer Vision and
  Pattern Recognition}, pages 3456--3465, 2016.

\bibitem{qu2011microblogging}
Y.~Qu, C.~Huang, P.~Zhang, and J.~Zhang.
\newblock Microblogging after a major disaster in china: a case study of the
  2010 yushu earthquake.
\newblock In {\em Proceedings of the ACM 2011 conference on Computer supported
  cooperative work}. ACM, 2011.

\bibitem{ray2019keyphrase}
J.~Ray~Chowdhury, C.~Caragea, and D.~Caragea.
\newblock Keyphrase extraction from disaster-related tweets.
\newblock In {\em The World Wide Web Conference}, pages 1555--1566. ACM, 2019.

\bibitem{reese2016we}
A.~Reese.
\newblock How we'll predict the next natural disaster: Advances in natural
  hazard forecasting could help keep more people out of harm’s way.
\newblock {\em Discover Magazine, Sep}, 2016.

\bibitem{reuter2018fifteen}
C.~Reuter and M.-A. Kaufhold.
\newblock Fifteen years of social media in emergencies: a retrospective review
  and future directions for crisis informatics.
\newblock {\em Journal of Contingencies and Crisis Management}, 26(1):41--57,
  2018.

\bibitem{saif2012semantic}
H.~Saif, Y.~He, and H.~Alani.
\newblock Semantic sentiment analysis of twitter.
\newblock In {\em International semantic web conference}, pages 508--524.
  Springer, 2012.

\bibitem{sakaki2010earthquake}
T.~Sakaki, M.~Okazaki, and Y.~Matsuo.
\newblock Earthquake shakes twitter users: real-time event detection by social
  sensors.
\newblock In {\em Proceedings of the 19th international conference on World
  wide web}, pages 851--860. ACM, 2010.

\bibitem{sarma2018domain}
P.~K. Sarma, Y.~Liang, and W.~A. Sethares.
\newblock Domain adapted word embeddings for improved sentiment classification.
\newblock {\em arXiv preprint arXiv:1805.04576}, 2018.

\bibitem{tompson2014joint}
J.~J. Tompson, A.~Jain, Y.~LeCun, and C.~Bregler.
\newblock Joint training of a convolutional network and a graphical model for
  human pose estimation.
\newblock In {\em Advances in neural information processing systems}, pages
  1799--1807, 2014.

\bibitem{varga2013aid}
I.~Varga, M.~Sano, K.~Torisawa, C.~Hashimoto, K.~Ohtake, T.~Kawai, J.-H. Oh,
  and S.~De~Saeger.
\newblock Aid is out there: Looking for help from tweets during a large scale
  disaster.
\newblock In {\em Proceedings of the 51st Annual Meeting of the Association for
  Computational Linguistics}, 2013.

\bibitem{wang2017using}
C.-K. Wang, O.~Singh, Z.-L. Tang, and H.-J. Dai.
\newblock Using a recurrent neural network model for classification of tweets
  conveyed influenza-related information.
\newblock In {\em Proceedings of the International Workshop on Digital Disease
  Detection using Social Media 2017 (DDDSM-2017)}, pages 33--38, 2017.

\bibitem{wang2014concept}
F.~Wang, Z.~Wang, Z.~Li, and J.-R. Wen.
\newblock Concept-based short text classification and ranking.
\newblock In {\em Proceedings of the 23rd ACM International Conference on
  Conference on Information and Knowledge Management}, pages 1069--1078. ACM,
  2014.

\bibitem{wang2017combining}
J.~Wang, Z.~Wang, D.~Zhang, and J.~Yan.
\newblock Combining knowledge with deep convolutional neural networks for short
  text classification.
\newblock In {\em IJCAI}, pages 2915--2921, 2017.

\bibitem{zahera2019jointly}
H.~M. Zahera, M.~A. Sherif, and A.-C. Ngonga~Ngomo.
\newblock Jointly learning from social media and environmental data for typhoon
  intensity prediction.
\newblock In {\em Proceedings of the 10th International Conference on Knowledge
  Capture}, pages 231--234. ACM, 2019.

\bibitem{zheng2017joint}
L.~Zheng, V.~Noroozi, and P.~S. Yu.
\newblock Joint deep modeling of users and items using reviews for
  recommendation.
\newblock In {\em Proceedings of the Tenth ACM International Conference on Web
  Search and Data Mining}, pages 425--434. ACM, 2017.

\end{thebibliography}
\end{document}